# Using Kernel SHAP XAI Method to optimize the Network Anomaly Detection Model

**Khushnaseeb Roshan**
Department of Computer Science
Aligarh Muslim University
Aligarh, India
khushiasa@gmail.com

**Aasim Zafar**
Department of Computer Science
Aligarh Muslim University
Aligarh, India
azafar.cs@amu.ac.in

*Abstract*—Anomaly detection and its explanation is important in many research areas such as intrusion detection, fraud detection, unknown attack detection in network traffic and logs. It is challenging to identify the cause or explanation of "why one instance is an anomaly?" and the other is not due to its unbounded and lack of supervisory nature. The answer to this question is possible with the emerging technique of explainable artificial intelligence (XAI). XAI provides tools and techniques to interpret and explain the output and working of complex models such as Deep Learning (DL). This paper aims to detect and explain network anomalies with XAI, kernelSHAP method. The same approach is used to improve the network anomaly detection model in terms of accuracy, recall, precision and f-score. The experiment is conduced with the latest CICIDS2017 dataset. Two models are created (Model_1 and OPT_Model ) and compared. The overall accuracy and F-score of OPT_Model (when trained in unsupervised way) are 0.90 and 0.76, respectively.

*Keywords—Explainable AI; Autoencoder; Shapley Additive Explanation; Network Anomaly; Network Security.*

## I. INTRODUCTION

Anomaly detection based on Machine Learning (ML) and Deep Learning (DL) is an active research area in many domains such as fraud detection [1], anomaly-based intrusion detection [2], network anomaly detection [3] and much more. Network anomalies are the unknown pattern of interest in network traffic and logs. Due to the lack of supervisory information and its unbounded nature, it is challenging to detect network anomalies and their explanation. DL models are giving tremendous results in anomaly detection but are still criticized due to their back-box nature and lack of interpretation of their outputs. The researchers have proposed so many approaches to interpret and explain the output of complex models (e.g. DL based models) over the years [4] [5]. The purpose of this study focuses on so-called "unsupervised-anomaly detection" as well as its interpretation in the area of computer network traffic and logs. In the unsupervised DL algorithm, the autoencoder is used for this experimentation. Based on the autoencoder reconstruction error (RE); the normal and attack instance are separated. RE can explain anomalies but only up to some extent [6] [7]. Hence, the kernelSHAP, a model agnostic approach [8], is used to explain anomalies with shapley values. Shapley values provides the true contribution of each feature based on the RE.

The motivation of this work is driven by renewed attention in the field of explainable AI (XAI). XAI provide various method and tools to convert the black-box model into transparent, accountable and interpretable models [9]. If we understand the working of the complex ML and DL models, we can improve and explain its results up to some extent.

The contribution of the proposed work are as follows:

- A novel approach for selecting the best subset of features with shapley values without using the target class label is proposed.
- Based on the attack instances, the shapley values are computed for each feature, providing the true contribution to the RE.
- We selected the top features responsible for the anomalous behaviour of the attack instance and used only these features to build an improved model for network anomaly detection.
- kernelSHAP method, which is a model agnostic XAI approach and the latest CICIDS2017 dataset is used for experimentation.

This work also illustrates how XAI techniques can be used to improve the performance of complex models (DL models), as we did in this paper. The overall organization of the paper are as follows. Section II describes XAI. Section 0 includes related work. Section IV discusses the proposed approach, dataset and models. Section V discusses implementation and results. Finally, Section VI concludes this work.

## II. EXPLANABLE ARTIFICIAL INTELLIGENCE

Understandability, Comprehensibility, Interpretability, Explainability, Transparency etc., are the nomenclature used by the XAI research community [9] [10], as shown in Fig. 1. The term XAI is new, but the problem of explainability existed when the researcher's studied explaining the output and decision making procedure of the expert system [11]. This section focuses on various definitions of XAI (what ?), the need for XAI (why ?), various approaches related to XAI in brief (how?).

There is no standard and globally accepted definition of XAI; different authors quote XAI differently. In [12], DARPA defined XAI as "produce more explainable models, while maintaining a high level of learning performance (prediction accuracy); and enable human users to understand, appropriately, trust, and effectively manage the emerging generation of artificially intelligent partners". In the Cambridge Dictionary of English Language [13], the revoked definition is as follows "Given a certain audience, explainability refers to the details and reasons a model gives to make its functioning clear or easy to understand." Another definition of XAI by the organizer of xML Challenge [14] is "an innovation towards opening up the black-box of ML" and as "a challenge to create models and techniques that both accurate and provide a good, trustworthy explanation that will satisfy customer's needs".In general, the goal of the XAI is centred around generating more transparent, responsible and accountable models without compromising their performance (prediction accuracy).

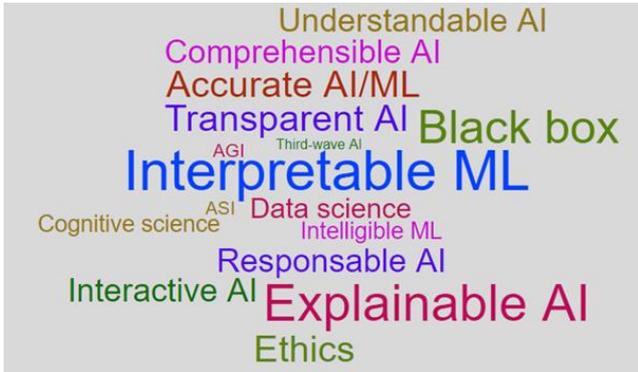

Fig. 1. XAI Word Cloud [9]

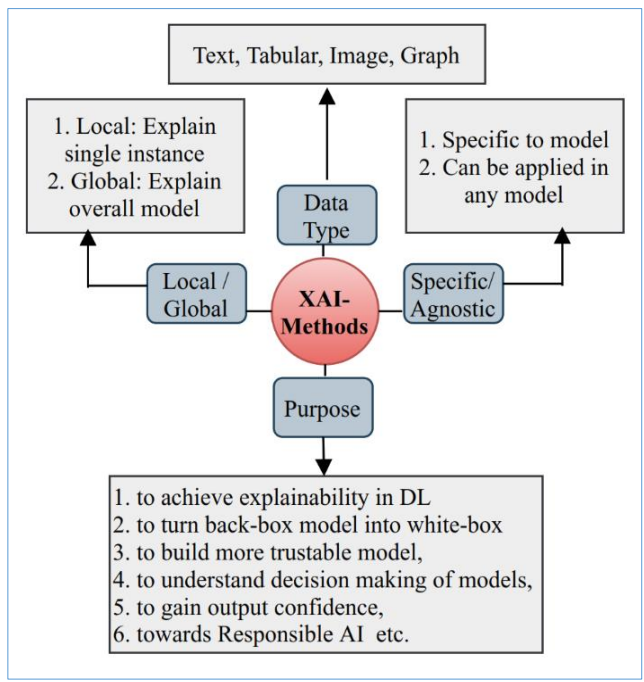

Fig. 2. Summarized mind map of XAI techniques

The need for XAI becomes clear at the points where the decision-making process of the models needs to be explained, such as in life-changing decisions (medical diagnosis) [15], in big financial decisions, in criminal justification etc. [9]. The explainability of the complex models (DL models) gives the control, discover and improve the results of the black-box models. And this will further remove the barrier of AI in real-life applications where the explanation and the interpretation of the output is the core aspect for its adoption. And the other key aspects related to the need for XAI are Trustworthiness, Causality, Transferability, Informativeness, Confidence, Fairness, Accessibility, Interactivity of DL, ML and ensemble learning-based models [10].

The different points of view and taxonomy of the XAI methods existed, such as the type of data (text, image etc.), methods related to global or local explanation. These methods can be model-specific, model agnostics, or both. Fig. 2 shows the summarized mind map of XAI methods, including their purposes. These are broader aspects and essential to be considered by researchers and practitioners while developing solutions in the field of XAI [16]. We would highly recommend some of the review and survey papers by the authors of [5] [9] [10] [16] for detailed analysis of various methods and application areas related to XAI.

### III. RELATED WORK

Computer Network anomaly detection is a wide research domain. So many approaches based on ML and DL has been proposed [17]–[23] by the research community in this domain. However, not much work has been done that detect and explain the cause of anomalous instances predicted by the models.

Antwarg et al. [24] used the autoencoder and kernelSHAP [8] to explain top anomalies detected by the model based on the RE. The model is trained in an unsupervised manner on benign data only (without target class feature). Both the real-world dataset such as KDDCUP 99 and the artificial datasets are used for evaluation. A similar approach is also shown in [6] by Takeishi. The author used the kernelSHAP method to explain anomalies based on the RE with PCA and shapley values. The study revealed that shapley value provided the true contribution of features in explaining the anomalies. Goodall et al. [25] presented situ a scalable solution for network attack/anomaly detection with intuitive visualization capability. Situ can monitor and discover suspicious behaviour within the network traffic logs and explain why the instance is anomalous. A case study in fraud detection and explanation was done by Collaris et al. [26]. The study revealed that different XAI techniques provided different results, yet all are valid and useful.

Outliers are also considered interesting anomalies, and several authors proposed methods for outlier explanation. Liu et al. [27] proposed the framework to explain the outlier with the model agnostic approach. The interpretation is based on the three aspects, namely outlier score, abnormal features and outlier context. Micenková et al. [28] proposed an approach to detect and explain the outlier with subspace separability. The outlier results validation is provided by the subset of the features for each outlier where the points are well separable from the rest of the data.

### IV. PROPOSED APPROACH

#### A. Unsupervised Features Selection based on SHAP

In this subsection, a method of features selection is proposed without using the target class variable. These features are selected based on the RE evaluated on the attack

Using Kernel SHAP XAI Method to optimize the Network Anomaly Detection Model

background set. The purpose is to select only those features which are actually contributing or affecting the RE, either increasing or decreasing it but in large magnitude. Consequently, these features would be the cause of major deviation of the RE and are more important to classify attack instances with high accuracy and recall. The alternate approach of this method can be thought of as selecting only those features that have large raw errors. However, just by looking only at the raw feature error, one can not identify the cause of the anomaly. For example, a large error on one feature can stem the anomalous behaviour of other features [6]. Hence, shapley values would be the best solution to find the true contribution of the features in the RE for attack instances. Further, these top contributing features are selected to build the optimized version of the model named OPT_Model.

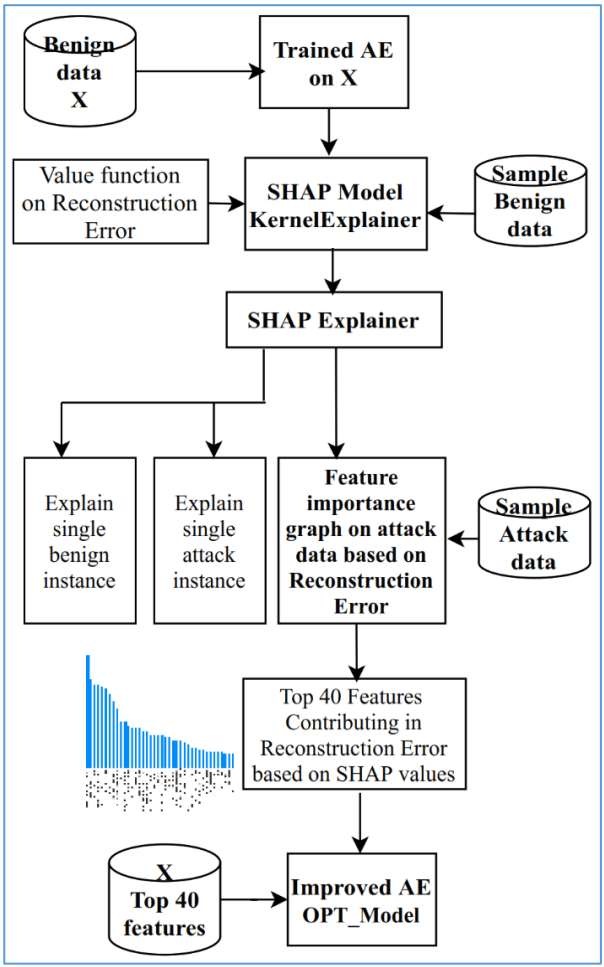

Fig. 3. Unsupervised feature selection approach with shapley values

The kernelSHAP [8] method is used to build the simple explanation model of the actual autoencoder model, i.e. Model_1. The kernelSHAP method is the model agnostic approach and requires access to the dataset and model's (Model_1) prediction function. But in our case, we used the autoencoder that simply reconstructs the original input, then how to define the prediction/value function for the explanation model? For this, the proposed approach by Takeishi [6] is used, as shown in Fig. 3. The author defined the valued function based on the reconstructed error $e(x)$ as in Eq. (1).

$$V(S) = \frac{1}{d} E_{p(x_{S^c}|x_S)}[e(x)] \qquad (1)$$

Here, $e(x)$ is the reconstruction error of the single test data instance $x \in R^d$ and $x$ is the concatenation of $x_{S^c} \in R^{d-|S|}$ and $x_S \in R^{|S|}$ as shown in Eq. (2). $S$ is the subset of features/indices within d and $S^c$ is the complement of $S$.

$$x = \begin{bmatrix} x_{S^c} \\ x_S \end{bmatrix} \qquad (2)$$

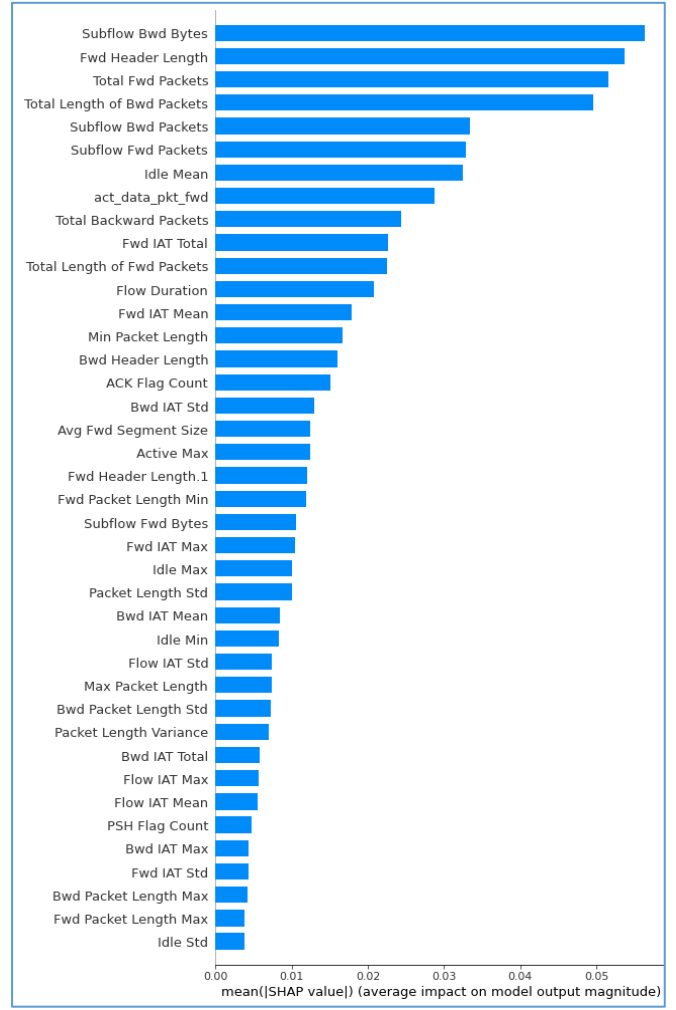

Fig. 4. Top forty features contributing to AE reconstruction error

The background set and the value function must be passed in kernelExplainer function to compute the features importance based on the RE of attack instances as shown in Fig. 3. In this case, the background set consists of 200 attack instances that are further processed with kmeans to improve its overall computation. Fig. 4 shows the top forty contributing features in RE of attack instances. And finally, these feature sets are used to build the optimized version of the model named OPT_Model.

*B. Autoencoder*

The autoencoder (AE) is an Unsupervised Artificial Neural Network (ANN) architecture first proposed by Rumelhart et al. [29]. The autoencoder consists of input layers, a number of hidden layers and output layers. Encoder and Decoder are two main components of the autoencoder. Encoder maps the input to the latent representation, and Decoder maps the latent representation back to the reconstructed output. The typical function used between the hidden layers is ReLU (rectified linear unit) [30]. The complete optimized architecture, such as the number of neurons in each layer, regularizer, learning rate, number of hidden layers etc., used in this experiment is shown in TABLE I.

TABLE I. PARAMETERS OF OPT_MODEL

| Parameters | Values |
|---|---|
| Layered Architecture | (50,20,8,20,50) |
| Epoch | 100 |
| Learning rate | 0.001 |
| Batch size | 8192 |
| Activation function | Relu |
| Loss function | MSE |
| Dataset split | 67 % - 33 % |

Furthermore, the Mean Squared Error (MSE), which is the most common function to measure the RE of the autoencoder, is used in this experiment. RE is the difference between the input and the reconstructed output as in Eq (3). In general, the RE is less on the benign data on which the autoencoder is trained; however, on the attack data, it is higher. The threshold is required to separate anomaly and normal data based on predicted MSE on the testing dataset by the models. For example, the instances on which the MSE is less than the threshold are labelled as benign (0) else anomaly (1).

$$MSE = \frac{1}{N}\sum_{i=1}^{N}(\bar{x} - x)^2 \qquad (3)$$

*C. Evaluation Metrics*

The accuracy is a widely used performance metric for ML models, but it is not always a good solution for the imbalance dataset. Hence both the models are evaluated based on the classification report consisting of Recall, Precision, F1-Score and Accuracy. The binary classification is separated into four groups known as a confusion matrix.

- True Negative (TN): correctly prediction of the negative class.
- True Positive (TP): correctly prediction of the positive class.
- False Negative (FN): incorrectly prediction of the negative class.
- False Positive (FP): incorrectly prediction of the positive class.

The metrics (Eq. (4) to (9).) are based on the above groups.

$$\text{Accuracy (ACC)} = \frac{TP + TN}{TP + TN + FP + FN} \qquad (4)$$

$$\text{Precision (P)} = \frac{TP}{TP + FP} \qquad (5)$$

$$\text{Recall (R)} = \frac{TP}{TP + FN} \qquad (6)$$

$$\text{F-Score (F)} = \frac{2 \times R \times P}{R + P} \qquad (7)$$

$$\text{Specificity} = \frac{TN}{TN + FP} \qquad (8)$$

$$\text{G -means} = \sqrt{Recall \times Specificity} \qquad (9)$$

In addition, the Reciever Operating Characteristics (ROC) curve and G-means are also used to select the best threshold for final classification [31] [32]. The G-mean [33] is the square root of specificity and recall. G-mean is considered as an unbiased classification metric with an optimal threshold selected based on the ROC curve [34].

V. IMPLEMENTATION AND RESULTS

The experiment is conducted on GPU enables Google Collaboratory, Python 3.7 and Keras Deep Learning Library.

*A. CICIDS2017 Dataset and Preprocessing*

The Canadian Institute of Cybersecurity created the CICIDS 2017 dataset [35], and the purpose is to develop the latest and most realistic background network traffic. This dataset is available in both packet-based and flow-based formats. The complete dataset is split into eight different files named Monday to Friday, having different attack classes. Monday file contains only benign data, and other files contain both benign and attack data. And there are a total of fourteen various attack class labels such as DDoS, FTP-Patator, SSH-Patator, Web Attack Brute Force and so on. However, in this experiment, the subset of the complete CICIDS2017 dataset is used. TABLE II. shows the instance count of benign and attack class labels used in this experiment. Monday file is used as training and validation of the models, and Friday file is used as test data.

The preprocessing is also done to replace all the null and infinity values from the dataset. These value has been replaced with the mean value, and then the scikit-learn StandardScaler function is used for feature scaling [36]. There are a total of seventy-eight features of the CICIDS2017 dataset. Model_1 is based on all features, and Model 2 (OPT_Model) is based on the forty features selected by the kernelSHAP method.

TABLE II. SUBSET OF CICIDS2017 DATASET

| Category | Class | Instances Count |
|---|---|---|
| Training Data | Benign/Normal | 150000 |
| Testing Data | Benign/Normal | 50000 |
| | DDoS/Attack | 10000 |
| Total | | 210000 |
| SHAP Background set | DDoS/Attack | 200 |

*B. Instance Level Explanation*

The kernelExpaliner method can provide both the instance level (single instance) and overall model explanation. The overall model evaluation is already discussed and shown in Fig. 4. In **Error! Reference source not found.**, the single normal and attack instance are explained in a way that "what are the features causing the significant deviation in the RE ?". The red colour indicates the feature value causing an increase in RE; however blue colour indicates that the corresponding feature is decreasing its value.

Using Kernel SHAP XAI Method to optimize the Network Anomaly Detection Model

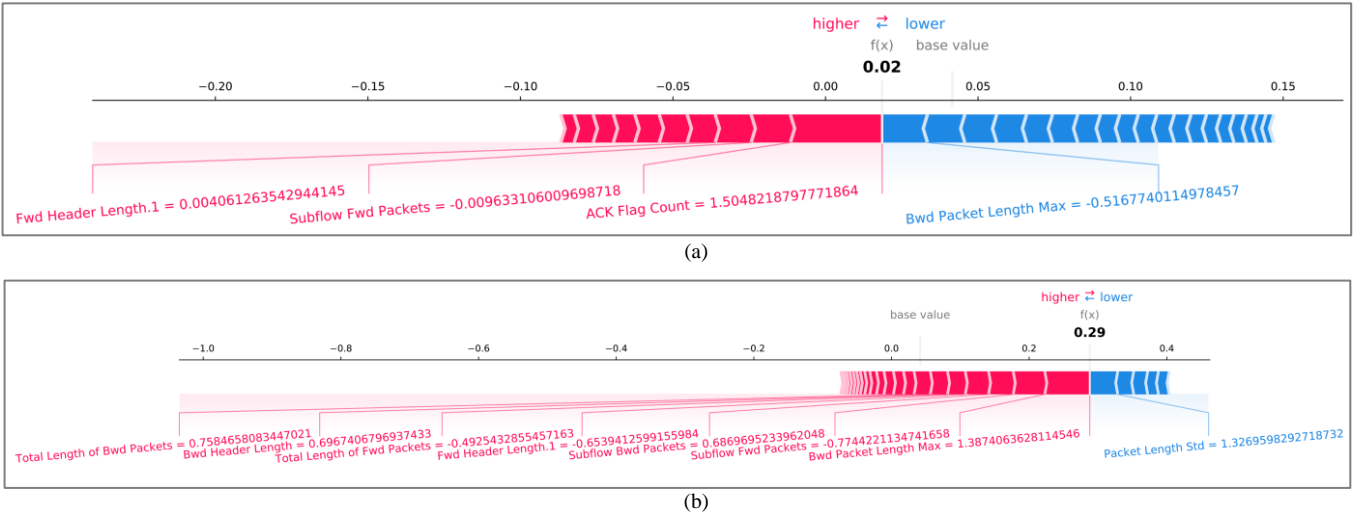

(a)

(b)

Fig. 5. (a) Features responsible for reconstuction error 0.02 for normal instance, (b) Features responsible for reconstruction error 0.29 for attack instance

In **Error! Reference source not found.** (a), the RE of the benign instance is 0.02, and the base value of the explainer model (kernelExplainer) lies between 0.0 to 0.5 in this case. The base value or the Expected value E[f(x)] is defined as "the value that would be predicted if we did not know any feature value for the current output f(x)" [8]. The features names which are causing major deviation in RE is also shown. Similarly, in **Error! Reference source not found.**(b), the attack instance with RE of 0.29 and features responsible for this deviation is also visible. This way, one can explain anomalies based on the features causing this large RE.

*C. Models' Comparision and Discussion*

Two models were built. Model_1 is one with all features set, and the second model is based on the selected forty features proposed in this paper. The top forty features which are highly contributing to the reconstruction error on the attack background set is selected based on the kernelExplainer function. These feature sets are used to train and validate OPT_Model. The architecture remains the same in the training and validation procedure for Model_1 and OPT_Model for comparison purposes. For both models, the MSE is predicted. Then based on the ROC curve in Fig. 6, the optimal threshold is selected to evaluate the confusion matrix and classification report, as shown in Fig. 7 and Fig. 8.

The G-mean based on the ROC curve, the optimal threshold for predicted MSE and area under the ROC (AUC) is shown in TABLE III. It is observed that the AUC for the OPT_Model is 0.95, which is higher than the Model_1 AUC, i.e. 0.804. AUC is a good measure for class imbalance datasets ad provides a single number (AUC) to compare different models [31] [37].

Finally, based on the threshold, the predicted MSE is converted into binary labels, i.e. "0" and "1". For example, the optimal threshold for Model_1 is 0.22, then the MSE greater than or equal to 0.22 is labelled as "1" otherwise as "0". Fig. 7 and Fig. 8 shows the comparison between Model_1 and OPT_Model based on the confusion matrix and classification report.

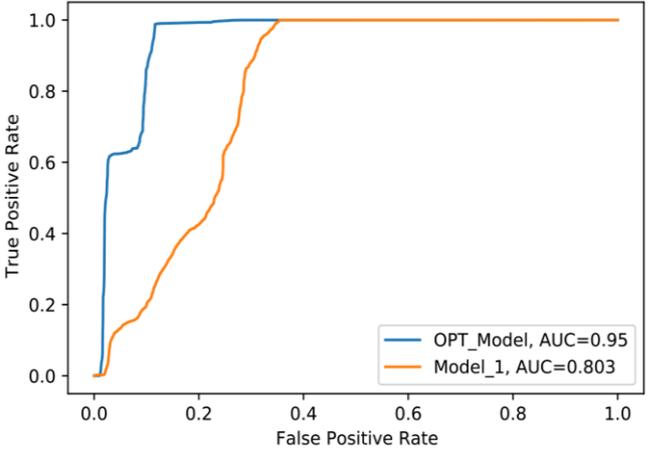

Fig. 6. ROC curves for Model_1 and OPT_Model

TABLE III. MODELS OUTPUTS

| Model | Features Count | AUC | G-mean | Optimal Threshold |
|---|---|---|---|---|
| Model_1 | 78 | 0.803 | 0.804 | 0.22 |
| OPT_Model | 40 | 0.95 | 0.934 | 0.54 |

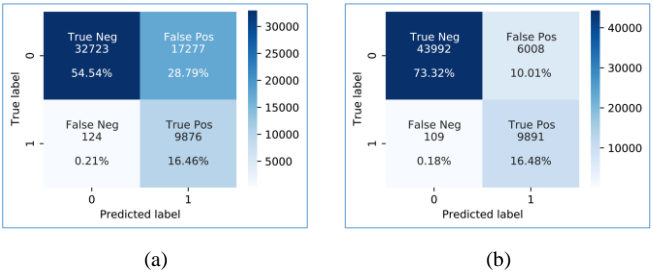

(a)  (b)

Fig. 7. Confusion matix for models (a) Model_1 and (b) OPT_Model

The novelty of the proposed work is how XAI can be used to improve the results of the DL models by selecting the most appropriate features in an unsupervised manner. Shapley values provides the true importannce of the features responsible for causing anomalous behaviour which clearly improved the performance of OPT_Model.

```
              precision    recall  f1-score   support

         0.0       1.00      0.65      0.79     50000
         1.0       0.36      0.99      0.53     10000

    accuracy                           0.71     60000
   macro avg       0.68      0.82      0.66     60000
weighted avg       0.89      0.71      0.75     60000
```
(a)

```
              precision    recall  f1-score   support

         0.0       1.00      0.88      0.93     50000
         1.0       0.62      0.99      0.76     10000

    accuracy                           0.90     60000
   macro avg       0.81      0.93      0.85     60000
weighted avg       0.93      0.90      0.91     60000
```
(b)

Fig. 8. Classification report for both models (a) Model_1 and (b) OPT_Model

The drawback of kenelSHAP method is its time complexity on the background set for complex and high dimensional datasets. If we increase the background set, the computation time will further increase or may take a couple of hours to compute shapley values for each feature. And selecting the appropriate background set from the complete dataset is also important and may affect the results of models [24].

VI. CONCLUSION

In this paper, we build the autoencoder based model that can detect and explain anomalies in the computer network traffic. Two models were created on the latest CICIDS2017 dataset. The first model is based on all features, and the second is based on selected features based on the kernelSHAP method, a model agnostic XAI approach. Top forty contributing features based on shapley values are selected to build OPT_Model, which outperformed the initial model (Model_1). This work also provides a brief introduction of the emerging XAI techniques in terms of "what ?", "why ?" and "how ?". XAI plays an important role in explaining and interpreting the results of complex models, such as DL models. And the models intrinsic (XAI) methods help in understanding the internal working of complex models.

The future extension of this work can be seen as applying other models agnostic or specific, global or local explanation techniques to explain and improve the results of unsupervised DL models for anomaly detection in the computer network traffic and other real-world applications as well.